%% file: root.tex
\documentclass[letterpaper, 10 pt, conference]{ieeeconf}  

\IEEEoverridecommandlockouts                              

\usepackage{amsmath}
\usepackage{graphics}
\usepackage{multirow}
\usepackage{svg}
\usepackage[ruled,vlined,linesnumbered]{algorithm2e}
\definecolor{mypink}{RGB}{255,0,252}
\definecolor{mygreen}{RGB}{21,219,15}
\definecolor{myorange}{RGB}{253,128,2}
\definecolor{mycyan}{RGB}{1,252,232}
\definecolor{originblue}{RGB}{68,114,196}
\usepackage{multirow}
\usepackage{array}
\usepackage{hyperref}

\AtBeginDocument{%
  \providecommand\BibTeX{{%
    \normalfont B\kern-0.5em{\scshape i\kern-0.25em b}\kern-0.8em\TeX}}}

\newcommand{\sssec}[1]{ {{\flushleft \textbf{#1}}}}

\newcommand{\sysname}{RadarHD\xspace}
\newcommand{\name}{RadarHD\xspace}
\newcommand{\newtext}[1]{{#1}}

\makeatletter
\newcommand*\bigcdot{\mathpalette\bigcdot@{2}}
\newcommand*\bigcdot@[2]{\mathbin{\vcenter{\hbox{\scalebox{#2}{$\m@th#1\bullet$}}}}}
\makeatother

\title{\LARGE \bf
High Resolution Point Clouds from mmWave Radar
}

\author{Akarsh Prabhakara, Tao Jin, Arnav Das, Gantavya Bhatt, Lilly Kumari, \\Elahe Soltanaghai, Jeff Bilmes, Swarun Kumar and Anthony Rowe%
\thanks{A. Prabhakara, T. Jin, S. Kumar and A. Rowe are with Carnegie Mellon University, Pittsburgh PA 15213 USA. Email: 
        {\tt\small \{aprabhak@andrew., taojin@andrew., swarun@, agr@ece.\} cmu.edu}}%
\thanks{A. Das, G. Bhatt, L. Kumari and J. Bilmes are with University of Washington, Seattle WA 98195 USA. Email:
        {\tt\small \{arnavmd2@, gbhatt@, lkumari@, bilmes@\} 
        uw.edu}}%
\thanks{E. Soltanaghai is with University of Illinois, Urbana-Champaign IL 61820 USA. Email:
        {\tt\small 
        elahe@illinois.edu}}%
\thanks{Corresponding Author: Akarsh Prabhakara.}%
}

\begin{document}

\maketitle
\thispagestyle{empty}
\pagestyle{empty}

\begin{abstract}

This paper explores a machine learning approach on data from a single-chip mmWave radar for generating high resolution point clouds -- a key sensing primitive for robotic applications such as mapping, odometry and localization. Unlike lidar and vision-based systems, mmWave radar can operate in harsh environments and see through occlusions like smoke, fog, and dust. Unfortunately, current mmWave processing techniques offer poor spatial resolution compared to lidar point clouds. This paper presents \sysname, an end-to-end neural network that constructs lidar-like point clouds from low resolution radar input. Enhancing radar images is challenging due to the presence of specular and spurious reflections. Radar data also doesn't map well to traditional image processing techniques due to the signal's sinc-like spreading pattern. We overcome these challenges by training \name\ on a large volume of raw I/Q radar data paired with lidar point clouds across diverse indoor settings. Our experiments show the ability to generate rich point clouds even in scenes unobserved during training and in the presence of heavy smoke occlusion. Further, \name's point clouds are high-quality enough to work with existing lidar odometry and mapping workflows.

\end{abstract}

\input{intro-2}
\input{related}

\input{network}
\input{implementation}

\input{evaluation}

\input{limitations}
\bibliographystyle{IEEEtran}
\bibliography{IEEEexample}

\end{document}

%% file: intro-2.tex
\section{Introduction} \label{sec:intro}

Lidar is often considered the gold standard in terms of sensors used for mapping, localization, and collision avoidance in robotics. A key enabler is its ability to generate low-noise and high-density point clouds, which can be easily tracked from one position to the next. 
Despite its ubiquitous use, lidar, just like visible light cameras, fail when used in environments with occlusions -- e.g. when robots navigate thick fog, smoke or dust, say for search-and-rescue, disaster recovery and firefighting. Such applications demand sensors that see past occlusions and sense the world in high fidelity.

Radars (the Radio Frequency - RF - equivalent of lidars) show promise given the robustness of RF waves to occlusions~\cite{starr2014evaluation}. However, due to the longer wavelengths of RF (even at mmWave frequencies), single-chip radars achieve an angular resolution that is two orders of magnitude (\newtext{hundred times}) lower than a lidar. Therefore, a radar resolves point clouds at a much lower resolution than lidar, limiting them to coarse-grained collision avoidance type applications. Higher resolution applications often resort to large mechanical radars, adding bulk and cost. Our goal is to push the resolution limits of a lightweight and compact, single-chip radar, suitable for much more portable platforms (e.g. future small robots, drones, AR/VR headsets, and mobile phones). We specifically seek to exploit the enormous amounts of low-level RF data normally discarded by traditional mmWave processing to dramatically enhance resolution. 


Current techniques to improve radar angular resolution include (1) synthetic aperture, which moves radars along specific trajectories precisely \cite{watts20162d,yanik2019near,prabhakara2020osprey,qian2020,guan2020through}, and (2) multi-modality (camera or lidar) sensor fusion for better information on angular resolution \cite{nabati2021centerfusion,long2021full,lekic2019automotive}. However, neither of these are applicable for radars that can move arbitrarily \newtext{or remain static}, and as previously stated, occlusions cause auxiliary sensors like cameras and lidars to fail. 

\newtext{We propose \sysname, which is a customized end-to-end neural network that generates \textit{lidar-like} point clouds from low resolution radar point clouds. We opt for an end-to-end learning-based pipeline to generate point clouds from radar, allowing for learning features ordinarily missed or thrown away by traditional signal processing pipelines. We show that our generated point clouds are excellent for scene capture, odometry, and mapping, even in smoke-filled environments.}

\newtext{\sysname\ is inspired by important recent work on using neural networks on mmWave radar for individual applications: odometry and mapping~\cite{milliego,lu2019millimap}. However, unlike these systems that target specific higher-level applications, \sysname\ targets a broader and more general problem: generating high resolution point cloud data directly from radar I/Q streams that is as good as what a lidar would output (albeit working in lidar-denied scenes). Our approach has two key benefits over per-application end-to-end learning: (1) Point clouds provide an interpretable, easy to understand output. For instance, it's more intuitive to debug and reason about point cloud errors rather than odometry errors, especially when both are output of a machine learning pipeline. (2) High quality point clouds enable a more general representation that can replace lidar in existing point cloud processing pipelines for several tasks beyond just odometry and mapping such as object detection and classification.}

\begin{figure}
\centering
\includegraphics[width=\columnwidth]{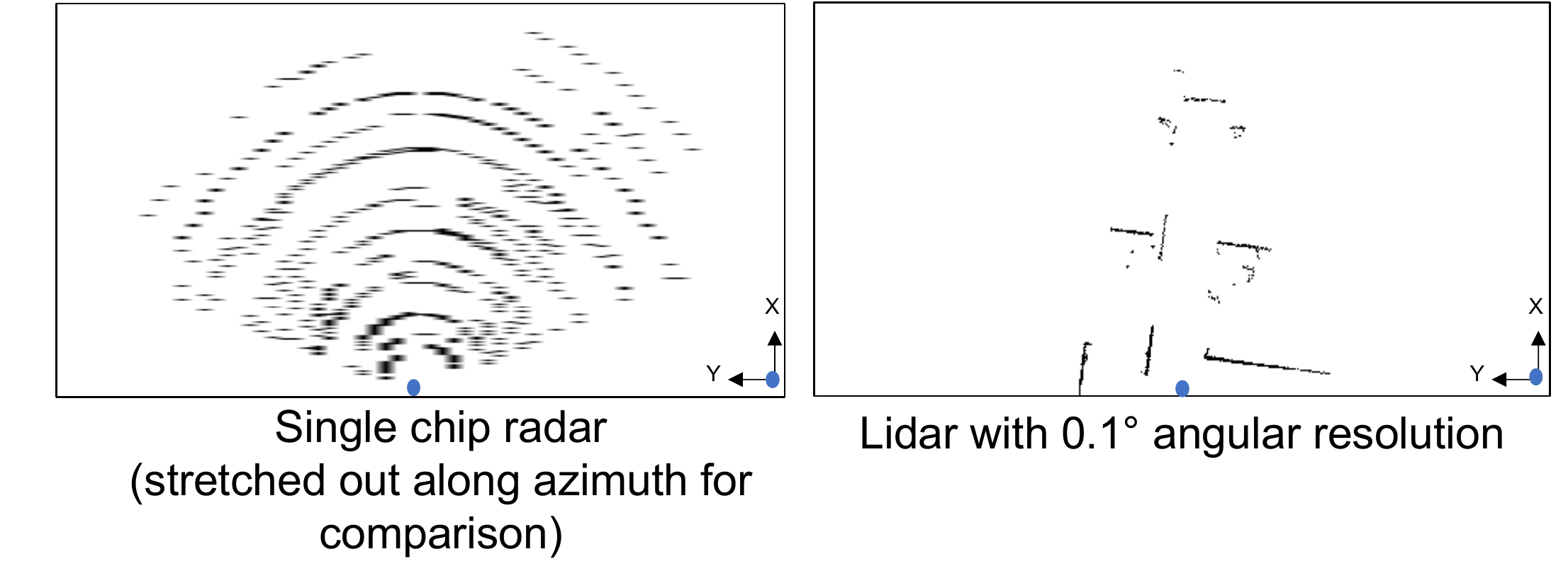}
\vspace*{-0.3in}
\caption{Single chip radar clearly has a much lower angular resolution than state-of-the-art lidar. Noticeably, radar also shows other spurious artifacts. \textcolor{originblue}{$\bigcdot$} marks the origin and both images show the same scene as perceived by each sensor in a 10x20m area.}
\vspace*{-0.25in}
\label{fig:2}
\end{figure}

In designing \name, we encountered two challenges. \textit{First}, raw radar measurements are impacted by various spurious artifacts -- sidelobes from strong reflectors that create sinc-like patterns across azimuth due to its poor azimuth resolution (see Fig. \ref{fig:2}), specular reflections from certain objects, and other processing artifacts \cite{bansal2020pointillism}. Eliminating these artifacts to recover the true objects is crucial for constructing a dense, accurate point cloud. \textit{Second}, radar images are coarse -- meaning that they struggle with resolving sharp environmental features with high angular resolution. In other words, the data from a low resolution radar is quite different from low resolution camera images, where na\"ive super resolution such as interpolation would give a sensible result.

\sysname\ overcomes these challenges by posing a supervised learning problem where large datasets of radar-lidar pairs collected on identical scenes, are used to inform radar to distinguish true objects from noise/artifacts. \name's core contribution is the customization of the entire neural network pipeline -- input/output representation, architecture and loss functions -- to tackle each of the challenges mentioned above.

To implement \name, we collect a large repository (200k pairs) of raw I/Q radar data from TI AWR1843 mmWave radars paired with lidar point clouds across different indoor and outdoor environments for generalization. \name's evaluation reveals low point-cloud error (24 cm) versus lidar ground-truth and 3.5$\times$ superior to traditional radar point-clouds. We also demonstrate the quality of our point clouds with two applications: odometry and mapping, using Google Cartographer~\cite{cartographer}.

\vspace*{0.05in} \noindent \textbf{Contributions: } \sysname makes three key contributions: 
\begin{itemize}
    \item Application of a super resolution model for generating lidar-like point clouds from  low resolution radar \footnote{Demo Link: \url{https://youtu.be/me8ozpgyy0M}}.
    \item A detailed evaluation of the system in new, unseen environments and severe occlusion such as smoke.
    \item A large repository of raw radar I/Q and lidar pairs along with source code.
\end{itemize}

%% file: related.tex
\section{Related Work} 


\sssec{Radar Super Resolution: } \newtext{The mm-level wavelength and the wide bandwidth available at mmWave frequency range provide high ranging accuracy and sensitivity. Combined with the robustness of mmWave radars to different lighting and weather conditions, mmWave radar is a popular option for sensing purposes \cite{guidi2015personal,lu2020see,lien2016soli,singh2019radhar,wei2015mtrack,zhao2018through,zhao2018rf,adib2015capturing,wu2020mmtrack}. In radar, high resolution is usually achieved by using Synthetic Aperture Radar (SAR) \cite{mamandipoor201460,yamada2017high,yanik2019near,prabhakara2020osprey,qian2020} or sensor fusion such as integrating radar and camera/lidar \cite{meng2018passive,ghasr2016wideband,appleby2007millimeter,sheen2007near}. While SAR is used in mobile contexts such as satellite imagery and automotive \cite{mostajabi2020high}, inaccurate motion information causes errors in the synthesized image \cite{fornaro2005motion}. For more portable applications (e.g: future small robots, light-weight UAVs etc.) that we envision, mm-accurate (on the scale of mmWave wavelength) motion information can be expensive to obtain. Moreover, our applications need high resolution images even when the radar is not being moved/temporarily static. More recently, techniques leverage deep learning \cite{guan2020through, fang2020superrf, geiss2020radar, sun2021deeppoint, cheng2022novel} to perform radar super resolution. \cite{cheng2022novel} uses deep learning to only keep robust points from the range-doppler spectrum. We instead tackle super resolution and seek to create \textit{lidar-like} point cloud which not only have true radar points but other synthetically generated points that boost the resolution. HawkEye \cite{guan2020through} is the closest related work to \name. However, it relies on input data obtained from mechanically scanning the mmWave radar on a large aperture slider to perform SAR \cite{guan2020through,guansupplementary}. Other works such as \cite{fang2020superrf, sun2021deeppoint} just like \cite{guan2020through} deal with static radar platform and single object setup. That is, a static radar is looking at a single object (static car/person) whose 3D point cloud is of interest. In such setups where radar platform and object are static, a ground truth can be obtained from a SAR system \cite{fang2020superrf}, and low resolution radar could be trained to generate SAR-like output. In contrast, we want to generate high resolution output just from a single-chip radar when (1) radar is static/arbitrarily moving and (2) in real, complicated environments. This calls for rethinking how we collect data, what our ground truth is and design the entire learning pipeline to deal with radar artifacts that show up due to real, complicated environments.}



\sssec{Radar Odometry and Mapping: } 
\newtext{State-of-the-art radar odometry techniques \cite{rapp2015fast,aldera2019fast,aldera2019could,cen2018precise,schuster2016landmark,almalioglu2020milli,vivet2013localization,schuster2016robust,scannapieco2018experimental,callmer2011radar,quist2016radar} mainly rely on scan matching, in which spatial features from the radar images are matched against previous scans or a pre-determined map. However, due to spurious reflections and artifacts in radar data, the accuracy of these methods significantly drop for low resolution radars. Follow-up research in this area relies on more complex modeling such as using velocity estimates from radar scans \cite{barron20053d,kellner2013instantaneous,kellner2014instantaneous,doer2020radar,park20213d, doer2021x}, the fusion of radar and IMU data based on Extended Kalman Filtering \cite{kramer2020radar,scannapieco2017ultralight,almalioglu2020milli,doer2020ekf}, or fusion of radar and RGB camera \cite{murdoch2021augmentation,mostafa2018radar,ouaknine2021carrada} to overcome radar limitations. Few other approaches involve end-to-end deep learning approaches for fusing Constant False Alarm Rate (CFAR) radar point clouds and IMU to obtain either odometry \cite{milliego} or mapping \cite{lu2019millimap, xulearned} individually. \name, on the other hand, seeks to solve an orthogonal problem: replacing CFAR points by creating higher resolution point clouds. These point clouds are general-purpose and can be fed into multiple lidar processing pipelines such as odometry, mapping, object detection, classification etc.}

\sssec{Radar-Lidar Datasets:} \newtext{Most publicly available radar and lidar datasets capture outdoor environments when mounted on a car \cite{RadarRobotCarDatasetICRA2020,meyer2019automotive,nuscenes2019}. Some of them use non-compact, bulkier mechanically scanning radar \cite{RadarRobotCarDatasetICRA2020}. We want a large repository of single-chip radar and lidar pairs in real, complicated environments for robotic applications. While there have been datasets for radar odometry \cite{milliego}, mapping \cite{lu2019millimap} and activity recognition \cite{singh2019radhar}, ground truth lidar point cloud is not commonly exposed. More recently, \cite{carrada,illinoisdatabank} provide single chip radar datasets with monocular and stereo camera largely for automotive purposes. \cite{kramer2021coloradar} is the closest that comes to the raw radar-lidar dataset that we desire. However, the indoor data samples ($\sim$15k radar-lidar pairs) are only a fraction of the entire dataset. To allow for larger volume of training data, custom design test cases and test robustly, we collect our own dataset.}

%% file: network.tex
\section{\textit{\sysname} System Design} \label{sec:arch}

\begin{figure}
\centering
\includegraphics[width=0.8\columnwidth]{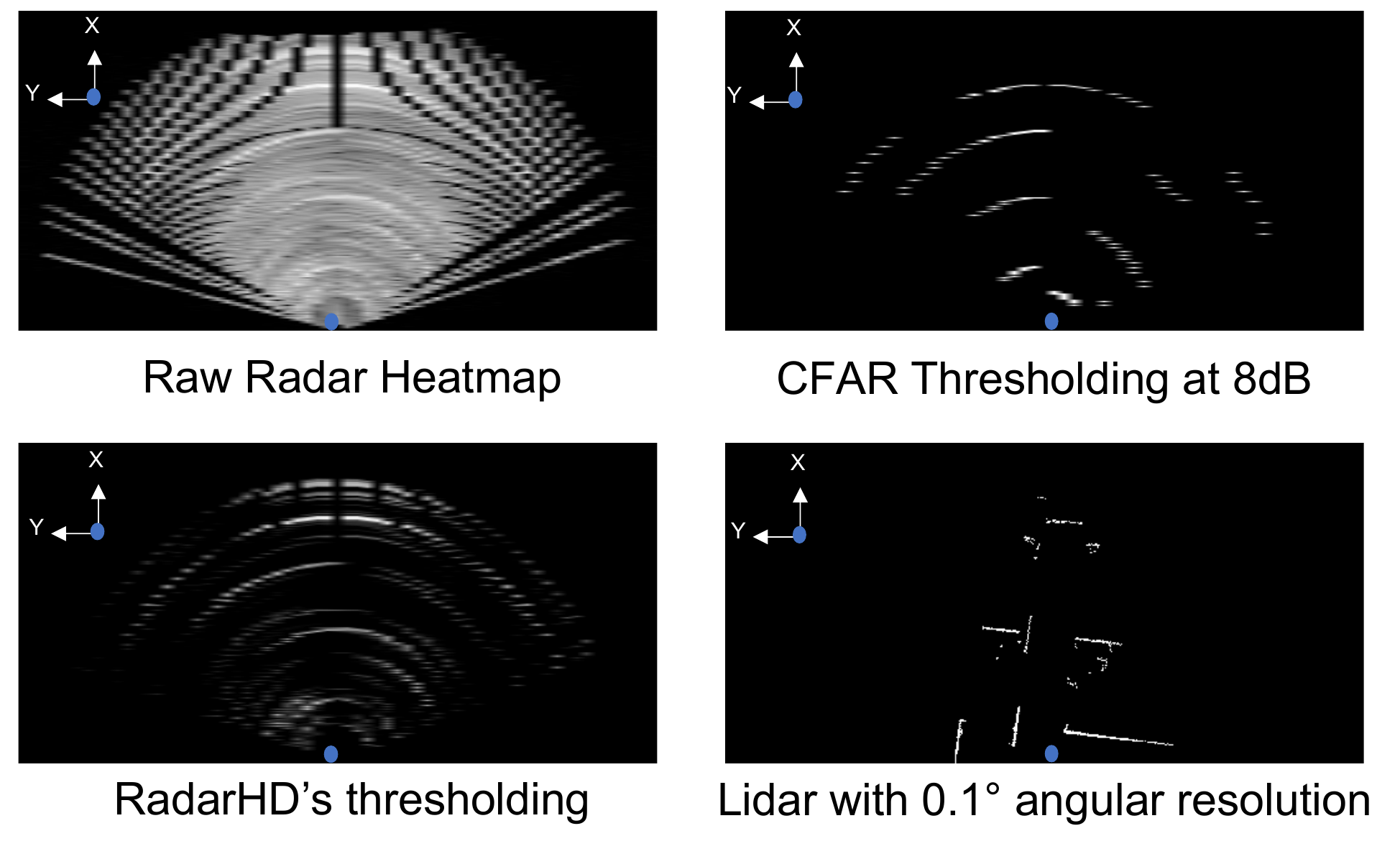}
\vspace*{-0.15in}
\caption{\newtext{Radar data representation: Raw radar heatmap captures all reflections and radar artifacts but the image is extremely crowded. CFAR thresholding only selects robust strong peaks. \name\ uses a hybrid capturing a combination of strong/weak peaks and radar artifacts. All radar images are log magnitude normalized to 8 bit image.}}
\vspace*{-0.25in}
\label{fig:cfar}
\end{figure}

The core objective of \sysname\ is to improve the resolution of the radar signals and make them \textit{lidar-like}. At first blush, one may consider simply interpolating the neighborhood and upscaling the radar image (Fig. \ref{fig:cfar}), much like super resolution on camera images \cite{smith1981bilinear}. But unlike camera images, low resolution radar images' neighborhoods lack similarity and can often be polluted. One such artifact affecting pixel neighborhoods is the azimuthally spreading patterns that can be seen in Fig. \ref{fig:cfar}. This spreading affects the entire image and is a result of side lobes originating from a few strong reflectors. Tackling this azimuth spreading requires us to go beyond local pixel neighborhood and get a global view of the radar image. On top of this, radar images have ``ghost points'', i.e. false detections of non-existent objects that also pollute pixel neighborhoods. Thus, we need to first understand what objects are truly present in the real world and weed out these artifacts before upscaling. \name\ goes beyond classical neighborhood approaches and explores machine learning approaches to obtain global image understanding.


\sssec{ML Design Choice: } Having motivated the need for learning, we now motivate \name's specific choice of ML models. One may simply consider a CNN model that uses convolution layers successively and builds a global view of the input image. However, since we are interested ultimately in the task of super resolution we need to consider global image understanding and upsampling in conjunction. Therefore, \name\ builds on a U-net based encoder-decoder architecture which is traditionally used on camera image data for segmentation and ports it to address the radar super resolution task. The U-net's general design allows the encoder to understand various radar artifacts and obtain semantically accurate representation of real world objects and the decoder uses this representation to do super resolution. 

\sssec{Design Challenges: } The rest of this section describes the key design decisions in \name's ML pipeline that helps U-net to learn effectively. (1) Effectively representing, pre-processing, and inputting radar I/Q data. (2) Designing U-net itself to allow for the elimination of spurious artifacts while preserving data from true objects in the scene. (3) Designing loss functions to ensure that features such as sharp lines in the expected output (see Fig. \ref{fig:cfar}) are preserved. 



\subsection{Radar Data Representation} \label{sec:pre}
We first consider the problem of representing radar signals as input to the ML architecture. To do so, we briefly describe our radar platform's capabilities and typical sensor output. 


\sssec{Radar Platform: } Our mmWave radar \cite{awr} provides raw I/Q data streams that can undergo further processing as needed. A typical radar processing pipeline \cite{iovescu2017fundamentals} involves a spatial Fourier transform that outputs a 2D heatmap with intensity of the reflected radar signal across range and azimuth. Traditionally, the heatmap gets thresholded to a ``point cloud''. \newtext{Constant False Alarm Rate (CFAR) is one such threshold detector that keeps robust objects and eliminate noisy ones. Fig. \ref{fig:cfar} shows CFAR thresholding applied to radar heatmap.}


\sssec{Choice of Input Representation: } At this point, we have a choice -- do we feed in raw I/Q inputs directly or should we send processed point cloud data. On one hand, inputting raw I/Q requires the model to understand and learn phase information and to learn some fairly obvious initial steps (e.g. a Fourier Transform). On the other hand, sending in a highly thresholded heatmap may filter out important information that values below the threshold carry, e.g. feeble objects masked by sinc lobes from stronger reflections.

\newtext{\name\ instead takes an approach in between these two extremes. Specifically, \name\ applies a very low threshold to the processed heatmap so that it preserves dominant reflectors, feeble ones and many artifacts. Our objective is to retain a significant portion of the heatmap, including feeble reflectors while leaving it to the ML model to learn and filter out spurious artifacts. \name\ chooses a threshold such that extremely weak points are omitted, but is still low enough to propagate radar's artifacts, strong and feeble reflectors. For context in Fig. \ref{fig:cfar}, CFAR thresholding has 110 non-zero pixels and \name's thresholding has 1606 non-zero pixels.}



\sssec{Polar vs Cartesian Representation: } We note that the thresholded image above is actually in polar format i.e. range-azimuth, but is shown in Fig. \ref{fig:cfar} in Cartesian format for easy understanding. Given that the output we desire for a point cloud is Cartesian, one may consider the Cartesian representation as the natural choice of representation.

\begin{figure}
\centering
\includegraphics[width=0.9\columnwidth]{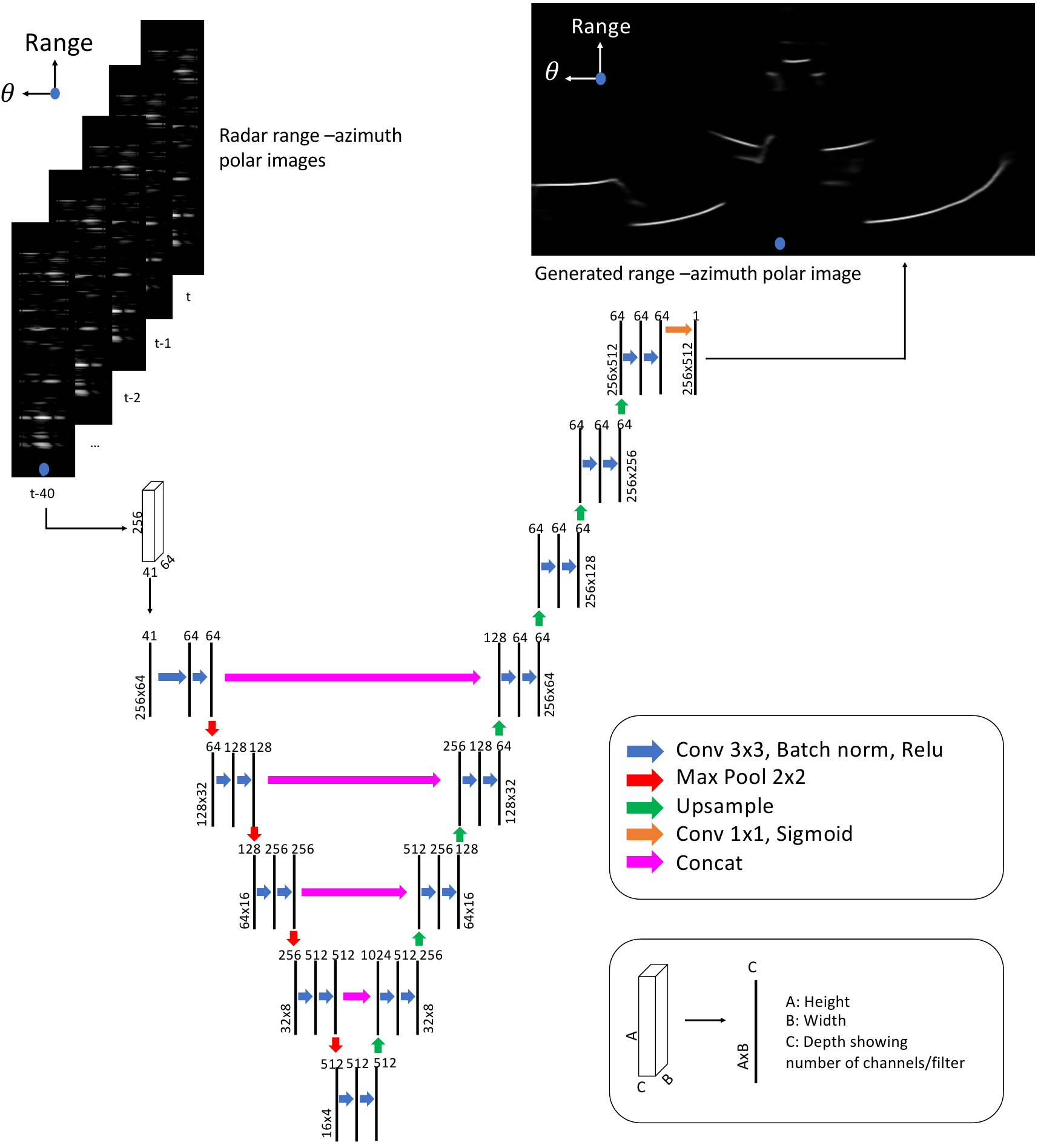}
\vspace*{-0.1in}
\caption{Assymetric U-net Network architecture: \sysname\ adapts an encoder-decoder architecture for radar super resolution. Colored arrows show different ML operations. Black solid lines are 3D inputs and outputs of these operations.}
\vspace*{-0.25in}
\label{fig:5}
\end{figure}

However, radar inherently measures radially and side lobes arising from a strong reflector also spread azimuthally. To capture these radial and azimuthal variations, one would need radial/azimuthal processing. But the primary learning element in convolutional layers in machine learning is a filter that performs 2D correlation across the height and width of the input. To leverage this to our advantage, we choose a polar format so that the filters then naturally traverse along the range and azimuth when they go across height and width, respectively. We thus have thresholded points on the heatmap with range, azimuth, and intensity arranged into an image with range (0-10 m) along rows and azimuth (-90$^{\circ}$ to $90^{\circ}$) along columns (Fig. \ref{fig:5}). The radar images ($64 \times 256$) are narrower than lidar ($512 \times 256$) because of the poorer azimuth resolution.

\subsection{Neural network architecture} \label{sec:nn}

We choose our base architecture as U-net \cite{ronneberger2015u} (see Fig. \ref{fig:5}) that allows encoder to denoise and obtain accurate semantic scene understanding and decoder to perform super resolution. U-net captures features at different ``resolution levels". From Fig. \ref{fig:5}, we can also clearly see that as we move to lower resolutions, the height and width of convolution output decreases (implying capturing more global understanding), and the number of filters increases (to capture richer global understanding). This is crucial to capture the azimuthally spreading artifacts that affects the radar image globally. \name's design has other salient differences from traditional U-nets for the context of radar super resolution.

\sssec{U-net Asymmetry due to Super Resolution: } While base U-net is generally a symmetric network, super resolution problem being fundamentally asymmetric (output image width larger than input), we adapt the base U-net to an asymmetric U-net by adding upsampling and convolutional layers to the decoder to obtain desired resolution (Fig. \ref{fig:5}).


\sssec{Mitigating Specularity: } While the input representation allows U-net to capture the azimuthally spreading artifacts, one key challenge that remains is ``ghost points''. This arises due to specularity \cite{bansal2020pointillism} of some radar reflections. Specularity is observed when an object, viewed from different orientations, appears and disappears in the radar image. One way to deal with this is to view the scene from multiple viewpoints. The need for multiple viewpoints is further exacerbated when we empirically observed the lack of persistence in the inferred images. That is, when the generated images are viewed as a video, objects would appear and disappear. Thus, it is important to consider using these multi-viewpoint images to tackle specularity and establish a notion of persistence.


Na\"ively one could use single frame inference and classical filtering to ensure that objects do not appear and disappear. We perform this filtering through the network by incorporating past radar frames (that offer multiple viewpoints if radar is moving) as input while performing super resolution on the current frame. Our design is to exploit the variable number of input image channels and stack the past frames as input channels to allow for understanding of each radar frame and modeling persistence across frames. We empirically analyze and find 40 past frames (2 seconds history) to be sufficient for enforcing persistence. Even in static cases, using past frames lead to smooth and less jittery output.


\subsection{Neural network training methodology}\label{sec:loss}
We carefully select loss functions to preserve features such as sharp lines that appear in lidar images (thin lines of white pixels against black background). We use a combination of various loss functions for achieving different objectives.


\begin{figure}
\centering
\includegraphics[width=0.77\columnwidth]{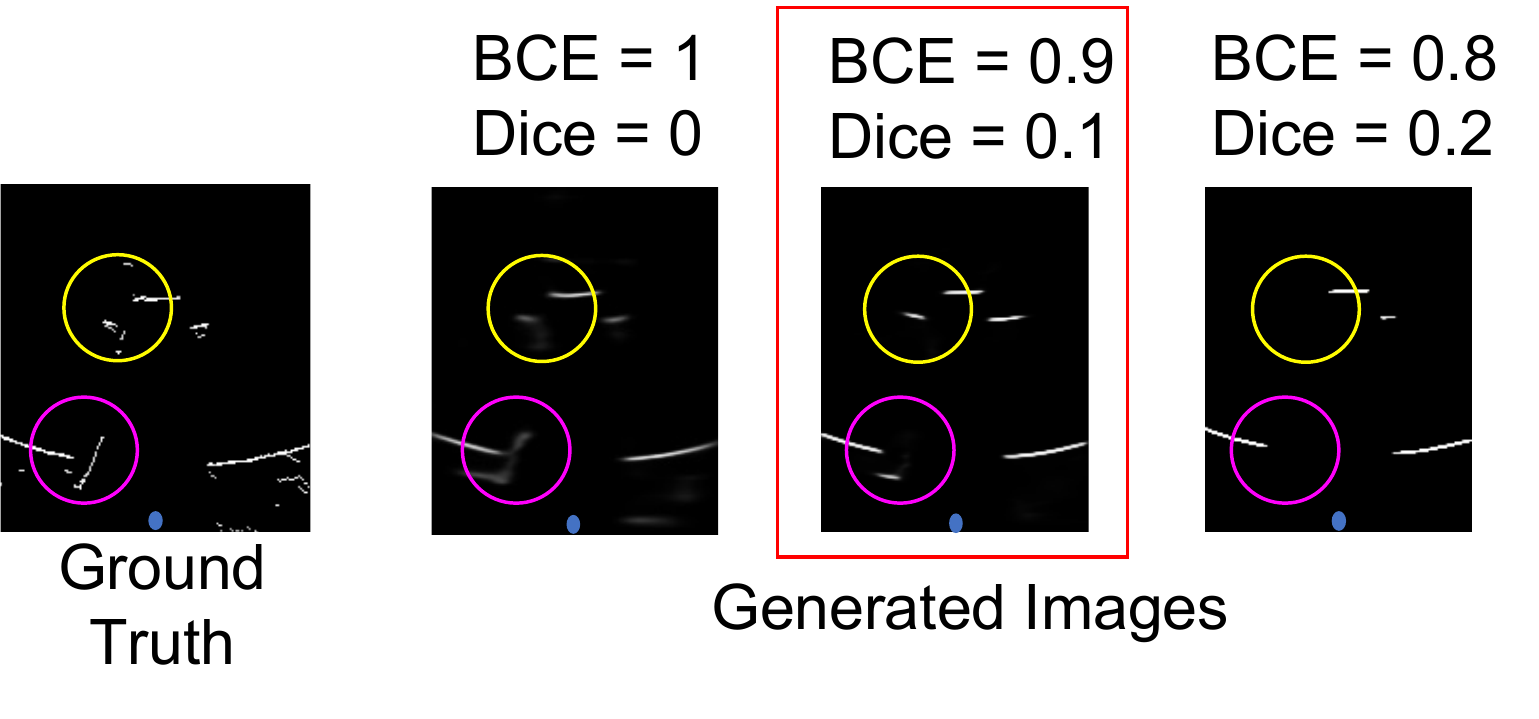}
\vspace*{-0.2in}
\caption{Qualitative Ablation Study: Dice Loss helps in sharpening lines, but is very aggressive and starts missing out features shown in colored circles. \name\ chooses a balance between sharpness and accuracy.}
\vspace*{-0.3in}
\label{fig:4}
\end{figure}


\sssec{Pixel wise loss:} To compare two images, one ground-truth label and one output from the network, we first consider the most standard loss function -- pixel wise loss. Our ground truth labels are binary lidar images. We compare this binary image against the final sigmoid layer output from the network. We use mean Binary Cross-Entropy (BCE) over all pixels. The objective of this pixel wise loss is to force each pixel to match the expected output.  Fig. \ref{fig:4} shows that BCE alone generates an acceptable output, but the lines and boundaries are not as sharp as the ground truth.

\begin{figure*}[!htb]
    \centering
    \begin{minipage}{0.23\textwidth}
        \centering
        \includegraphics[width=\textwidth,height=1.5in]{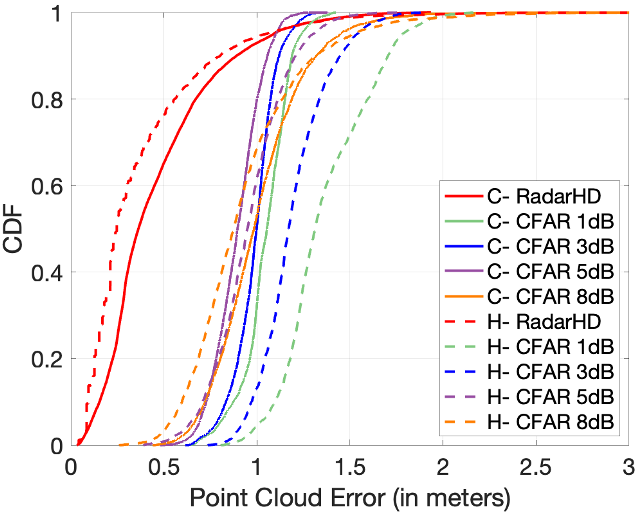}
        \vspace*{-0.2in}
        \caption{\name\ point cloud error. \\ \footnotesize{C- Chamfer H- Mod. Hausdorff}}
        \label{fig:cdf}
    \end{minipage}
    \hspace*{.05cm}
    \begin{minipage}{0.22\textwidth}
        \centering
        \includegraphics[width=\textwidth,height=1.5in]{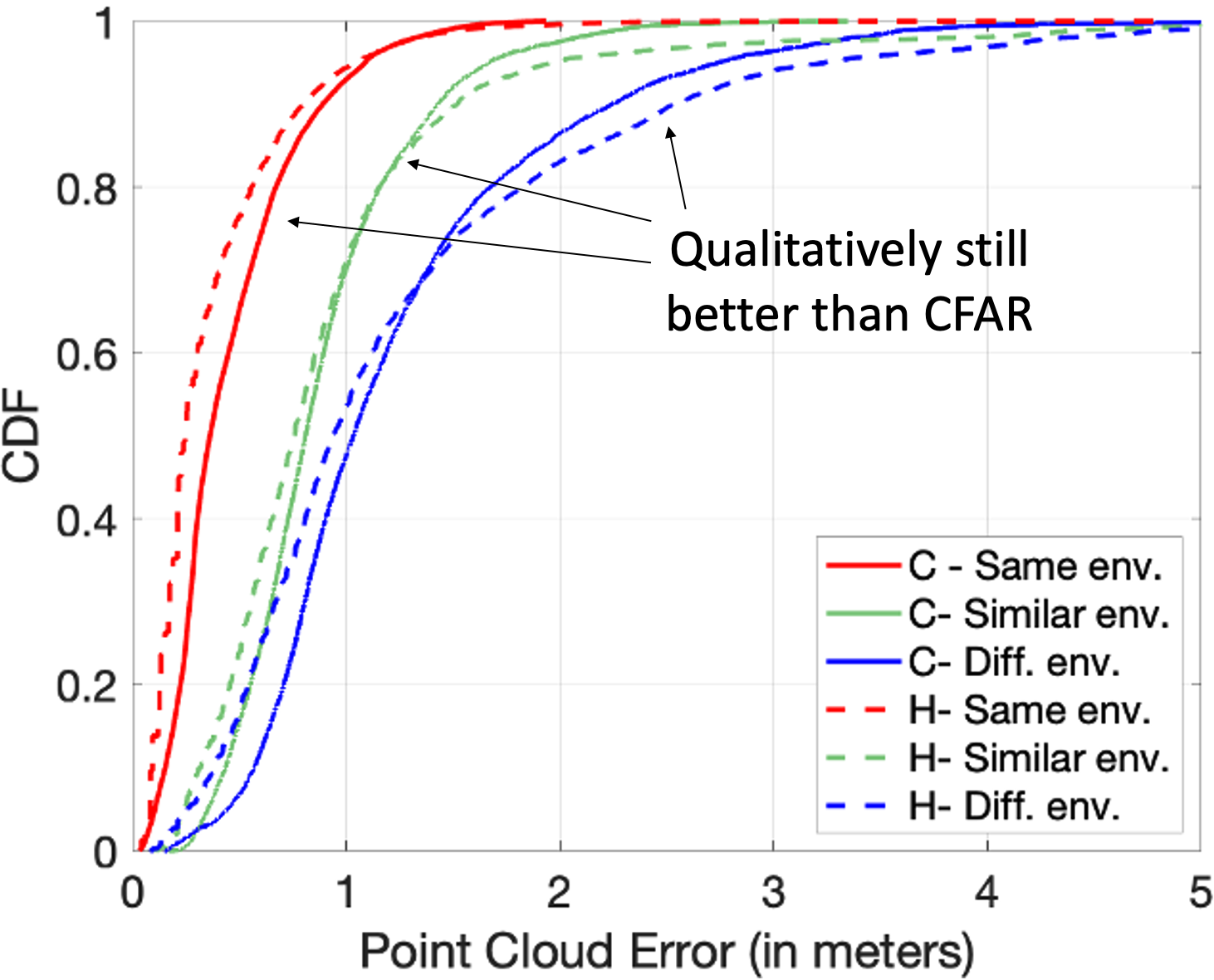}
        \vspace*{-0.2in}
        \caption{\newtext{Contrasting \name\ error in varying environments. \\\footnotesize{C- Chamfer H- Mod. Hausdorff}}}
        \label{fig:env}
    \end{minipage}%
    \hspace*{.05cm}
    \begin{minipage}{0.23\textwidth}
    \centering
    \includegraphics[width=\textwidth,height=1.5in]{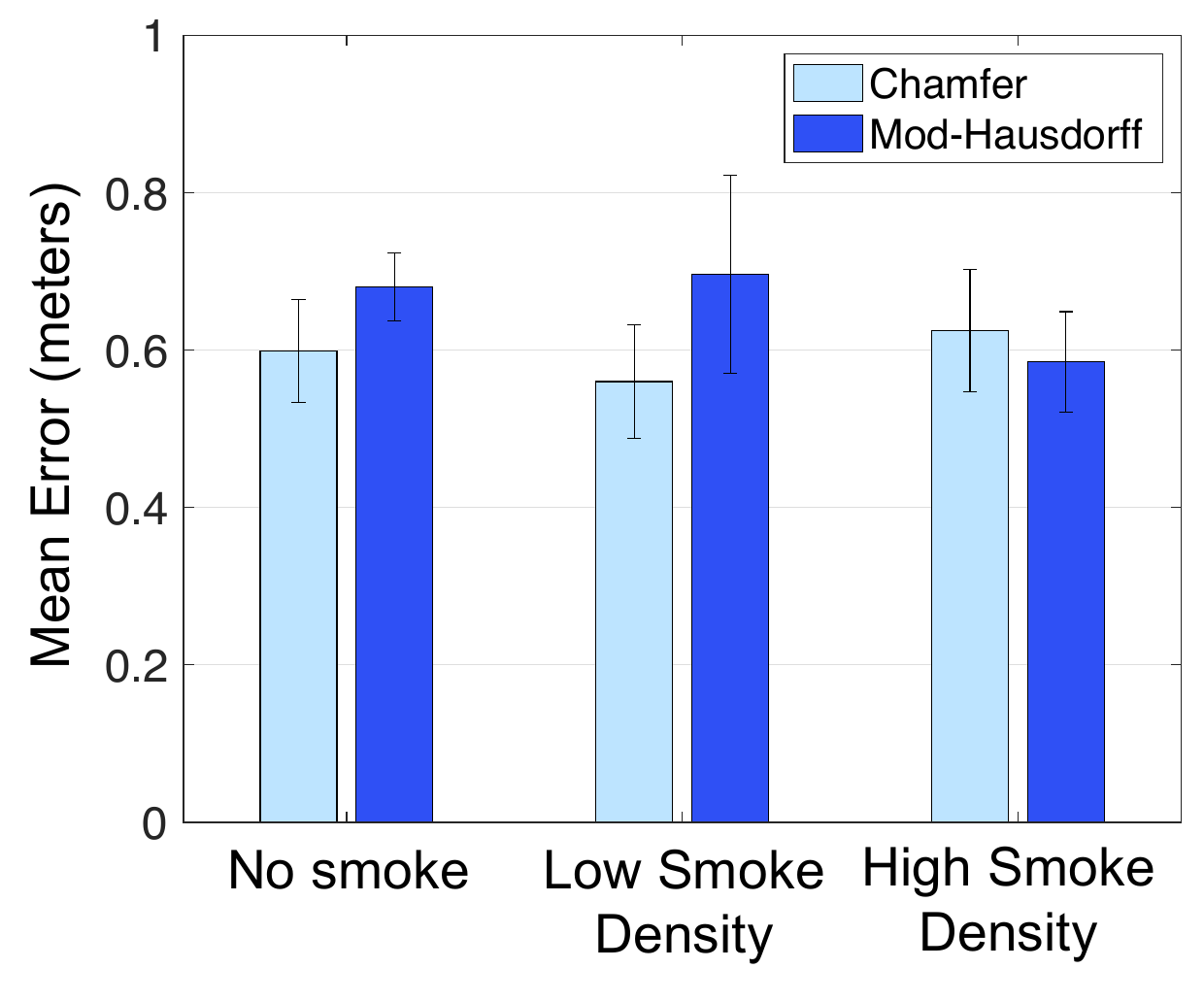}
    \vspace*{-0.2in}
    \caption{RadarHD is unaffected \\ in smoke.}
    \label{fig:smoke}
    \end{minipage}
    \hspace*{-.3cm}
    \begin{minipage}{0.3\textwidth}
    \centering
    \vspace*{0.4in}
    \includegraphics[width=\textwidth]{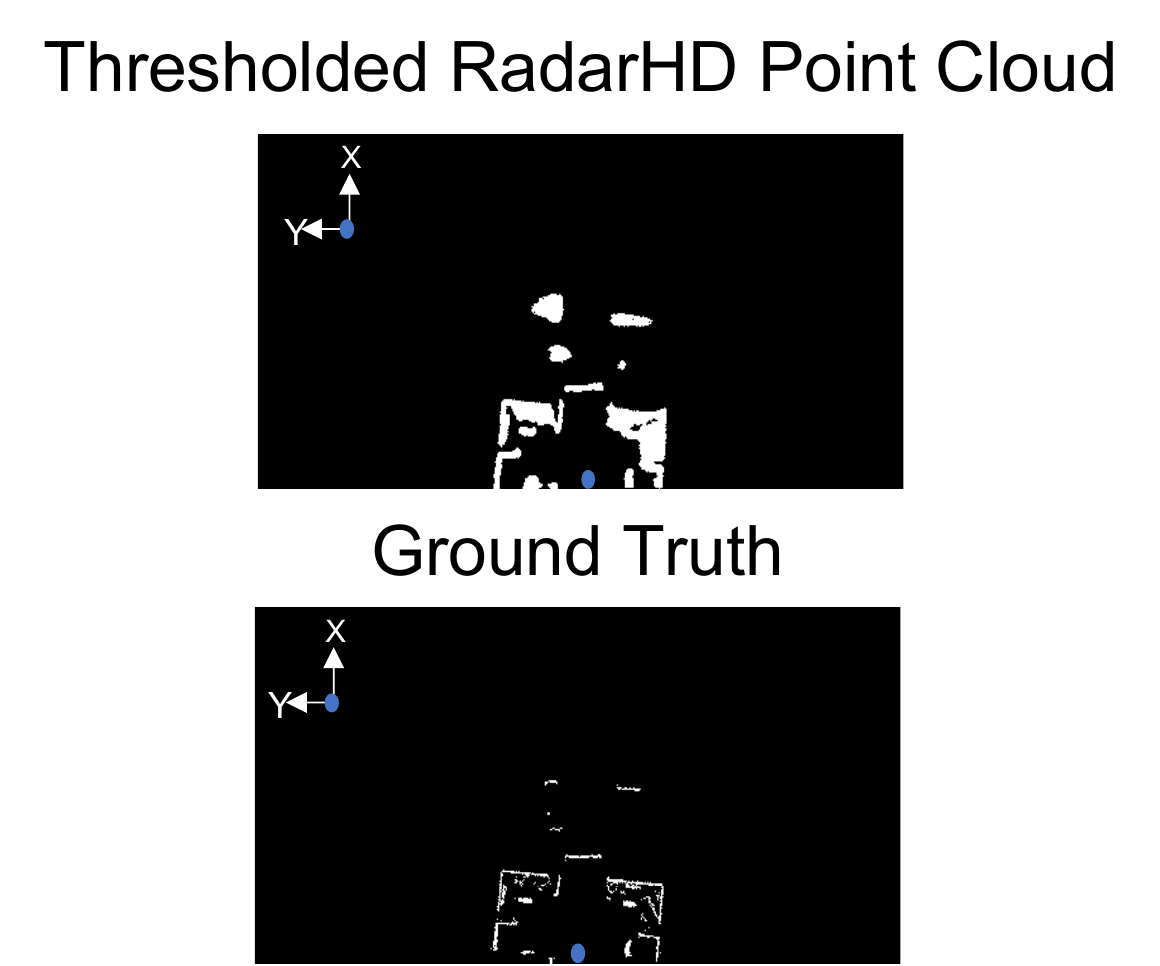}
    \vspace*{-0.2in}
    \caption{RadarHD thresholds generated images to compare with lidar. \textcolor{originblue}{$\bigcdot$} marks the origin and images show a 10x20m area. \\}
    \label{fig:example_pc}
    \end{minipage}%
    \vspace*{-0.4in}
\end{figure*}

\sssec{Dice loss:} To promote crisp and sharp lines in the output image, we draw from Dice loss \cite{dice} used in computer vision tasks like boundary detection \cite{deng2018learning}. For each pixel in ground-truth $g_i$ and network output $o_i$, Dice loss for $N$ pixels is: 
\begin{equation*}
D = 1 - \frac{2 \sum_{i=1}^{N} o_i g_i}{\sum_{i=1}^{N} o_i^{2} + \sum_{i=1}^{N} g_i^{2}}
\end{equation*}
Here, the numerator finds the loss pixel wise and is maximized when both $o$ and $g$ are identical. The denominator keeps a global view of total number of points that are 1. The loss promotes maximizing the intersection between $o$ and $g$ and penalizes the union of $o$ and $g$. This forces the network to output 1 exactly where the ground truth is 1, while remaining 0 where the ground truth is 0. This enables a sharper and crisper prediction than pixel wise cross entropy. We analyze and find trade-off between BCE and Dice loss (Fig. \ref{fig:4}). More Dice loss leads to eliminating certain important features.

%% file: implementation.tex
\section{Implementation} \label{sec:implement}


\begin{figure}
    \centering
    \vspace*{-0.18in}
    \includegraphics[width=0.95\columnwidth]{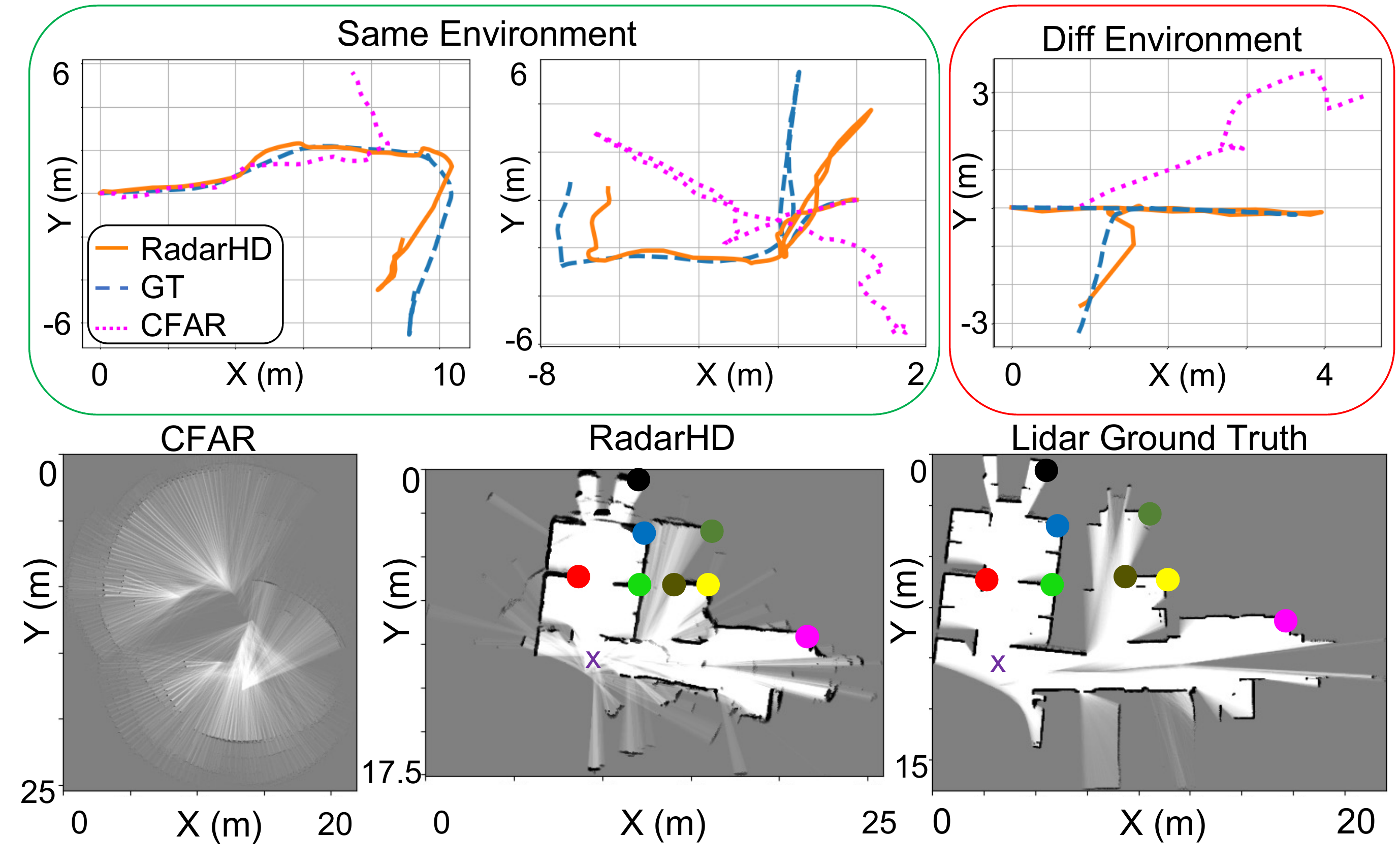}
    \vspace*{-0.15in}
    \caption{Odometry and Mapping Examples. Colored points in the maps show physically similar \textit{keypoints}. x is the start of trajectory.}
    \label{fig:traj_map_example}
    \vspace*{-0.2in}
\end{figure}

\sssec{System Hardware: } \name\ was implemented using TI mmWave radar AWR1843, a state-of-the-art single-chip radar with a theoretical range resolution of 3.75 cm and azimuth resolution of 15$^{\circ}$. \name's objective is to improve this azimuth resolution. We use the AWR1843 together with DCA1000EVM to collect raw I/Q samples.


\sssec{Testbed: } Our testbed consists of radar, lidar for ground truth and camera for debugging - all time-synced. Our testbed is mounted on a mobile testing rig. Our entire data repository consists of about 200k radar I/Q - lidar pairs collected across a total area of 5147 $m^{2}$ which we believe will be extremely useful to the research community.




\sssec{Ground Truth: } We use Ouster OS 0 - 64 beam mechanical scanning lidar for our ground truth. The lidar is configured to work at 0.35$^{\circ}$ azimuth resolution. We only use the forward-facing lidar points for super resolution and we also restrict the lidar's elevation FoV to be within +/-30cm. 


\sssec{Baselines: } \newtext{We use Constant False Alarm Rate (CFAR) based thresholding with different thresholds as our baselines. Today, CFAR is widely used for collecting radar point clouds. CFAR is an ideal baseline as: it is not machine learning based, relies purely on radar signals (no IMU) and works when radar is either static/mobile (unlike SAR). We specifically implement Cell-Averaging CFAR \cite{richards2005fundamentals}.}

\sssec{Train and Test Data: } We train our system in a large office space that includes furniture, electronics, walls, conference rooms, and cubicles, with a total area of 839 $m^{2}$. All of our tests are performed on unseen data in (1) \textit{same environment} with different trajectories collected across different days, (2) \textit{similar environment} in different office spaces with different cubicle structures, and (3) \textit{different environment}, which is totally new and includes building lobbies and outdoor environments. Across all our data we use 28 trajectories for training (561 meters long with 22784 image pairs in total) and 39 diverse trajectories for testing (714 meters long with 36779 image pairs in total) to allow for testing robustly.

%% file: evaluation.tex
\section{Results} \label{sec:evaluate}

\subsection{Point cloud Comparison}
\label{sec:pcc}

\sssec{Method:} After training on rich office space environment, we run \name\ on all the diverse test samples. To compare against lidar point cloud, we first convert the threshold the range-azimuth output image to obtain a list of points with their (x,y) location. We then compare point cloud error using two popularly used point cloud similarity metrics \cite{pdal_contributors}: (1) Chamfer distance \cite{chamfer}: finds the nearest neighbor for each point in one point cloud to the other, and takes the mean of all these distances to get an error for each point cloud pair. (2) Modified Hausdorff Distance \cite{hausdorff}: which also finds the nearest neighbors and obtains the median neighbor distance.

\sssec{Comparison to baseline:} Here, we show our performance in the floor-wide office environment on 19 different trajectories against different CFAR thresholds. This includes 18k radar-lidar point cloud pairs, each over a 10x20 meter area.

As seen in Fig. \ref{fig:cdf}, we obtain a 0.24 m modified-Hausdorff median error and 0.36 m Chamfer median error. CFAR, on the other hand, varies depending on the threshold. A low threshold like 1dB threshold creates point cloud 5x denser than lidar, while a high threshold like 8dB would just have 10\% of the number of points captured using lidar. Despite varying density levels across these extremes, none of them have any structural similarity to the ground truth lidar point cloud. So from 1dB to 8dB, as the threshold increases and density decreases, both point cloud error metric Cumulative Distribution Functions (CDFs) shift to the left. However, because the highest threshold generates a very sparse point cloud and does not resemble ground truth, the CDFs do not shift any more to the left with an increase in the threshold. In fact, beyond 8dB, the number of points was less than 1\% that of lidar. Our generated point cloud not only improves upon CFAR by 3.5$\times$ (mod-Hausdorff) and 2.7$\times$ (Chamfer), but is also structurally more similar to the ground truth as seen in Fig. \ref{fig:example_pc}, unlike Fig. \ref{fig:2}.



\sssec{Generalization:} We compare the performance of point cloud generation along these 19 diverse trajectories against those obtained in \textit{similar} and \textit{different} environments having trained only on \textit{same} environments. We use 7.5k point cloud pairs, each for \textit{similar} and \textit{different} environments. 

Fig. \ref{fig:env} shows the change in performance in new environments. We see a median error of 0.75 m (mod-Hausdorff) and 0.8 m (Chamfer) for \textit{similar} environments and 0.94 m (mod- Hausdorff) and 1.03 m (Chamfer) for \textit{different} environments. This tells us that these point clouds generated are not quite as accurate as when tested in \textit{same} environment. At first glance, we observe that these medians are similar to the medians for some CFAR CDFs in Fig. \ref{fig:cdf}. However, we would like to point out three important reasons why \name's point clouds are still superior. First, the CFAR CDFs start on the x-axis at 0.4 m; in contrast, even for \textit{similar} and \textit{different} environments, we see that the CDFs start at 0.08 m. This shows that there is a significant fraction of the point clouds that is accurately inferred by our system. Second, because both Chamfer distance and modified Hausdorff distance have a nearest neighbor point association, structural similarity is not entirely captured \cite{wu2021density}. However, we can qualitatively see from Fig. \ref{fig:example_pc} that our system indeed generates meaningful points. Third, to quantitatively show the impact of improved accuracy, we compare \name\ against CFAR in two key applications - odometry and mapping in Sec. \ref{sec:odometry}.

\begin{figure}
    \centering
    \includegraphics[width=0.7\columnwidth]{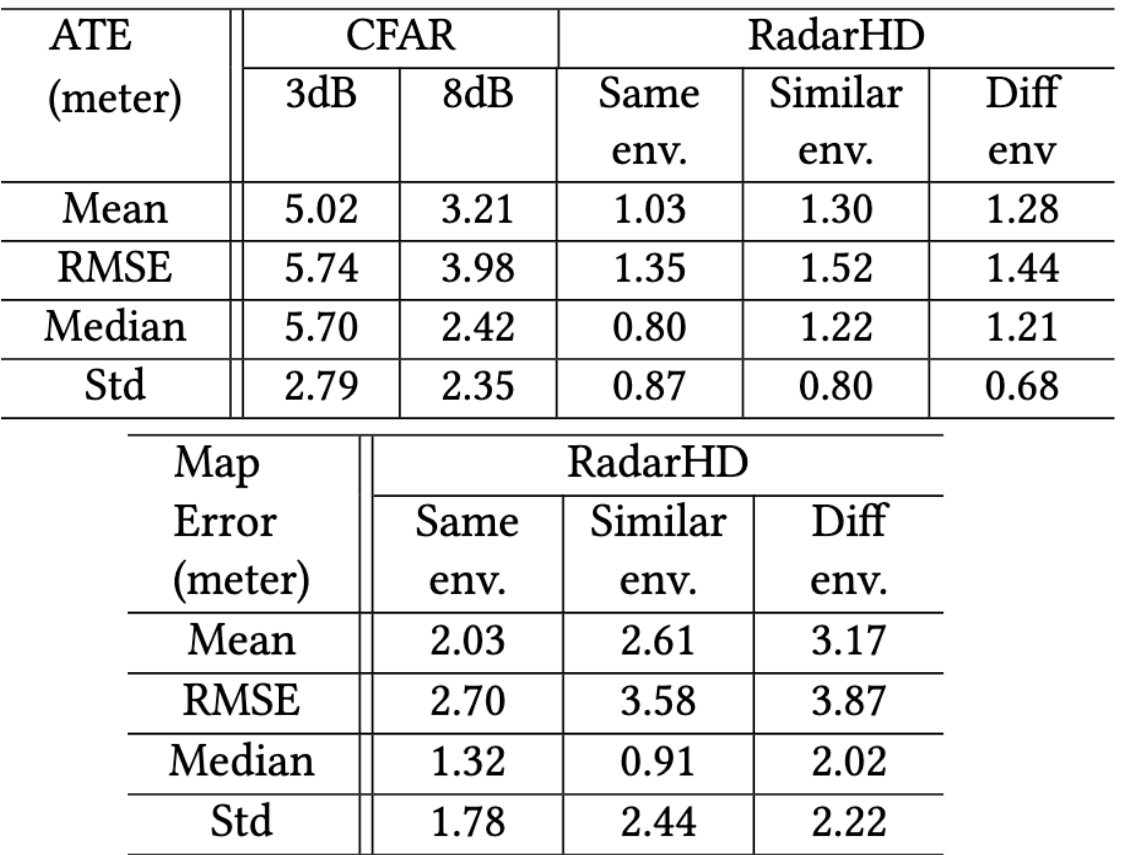}
    \vspace*{-0.1in}
    \caption{Odometry (Absolute Trajectory Error - ATE) and Mapping quantitative results.}
    \label{fig:traj_map_table}
    \vspace*{-0.3in}
\end{figure}

\sssec{Smoke:} To study the impact of occlusion, we build a smoke chamber around the testing rig and drop smoke pellets in it to create dense smoke. The radar is kept static and same scene is captured without smoke and varying levels of smoke. We use onboard camera to judge the intensity of smoke. 

We notice that even with 1 smoke pellet, that generates 500 cubic feet/min, lidar does not receive any points. However, the collected radar signals in smoke are almost identical to that without smoke even for the densest smoke we could create using 4 pellets. As the collected radar signal remains the same, we expect similar performance as without smoke. Fig. \ref{fig:smoke} validates this by showing that \name's performance doesn't degrade, up to densities we could create.

\subsection{Odometry and Mapping Comparison}\label{sec:odometry}
\sssec{Method:} Using the high quality point clouds generated by \name, we next show 2 downstream tasks that \name\ will enable in scenarios where lidars fail: odometry and mapping. Since our points are lidar-like, we evaluate this by feeding our point clouds, without adding any other sensor (e.g IMU), into existing lidar SLAM frameworks such as Google Cartographer \cite{cartographer}. We obtain the 3-DoF pose in 2D (translation (x,y) and rotation) and map from Cartographer.


\sssec{Odometry:} We evaluate odometry against lidar and benchmark CFAR point clouds using Absolute Trajectory Error - ATE (see Fig. \ref{fig:traj_map_table}). In all cases, including \textit{different} environments, the odometry accuracy of \name\ outperforms that of CFAR regardless of threshold. Qualitatively, one can clearly see the difference between \name\ odometry and CFAR odometry in Fig. \ref{fig:traj_map_example}. \name\ achieves performance comparable to $0.8$ m reported in radar+IMU pipelines dedicated for odometry in past work \cite{milliego}.

\sssec{Mapping:} We benchmark the mapping performance by identifying \textit{keypoints} that point to the same physical feature in the real world, such as corners of a room, and then calculate the Euclidean distance error of corresponding \textit{keypoints} between \name\ and ground truth. Fig. \ref{fig:traj_map_example} shows a qualitative comparison of a map generated from one trajectory. It is clear that CFAR does not provide any meaningful features to extract \textit{keypoints} while \name\ achieves a structurally similar map compared to lidar. Fig. \ref{fig:traj_map_table} shows the Euclidean distance error between \textit{keypoints} across different environments. Good performance on odometry/mapping is possible only because of artifact-free, meaningful point clouds generated by \name. \name\ also allows for visual debugging of point clouds in case of poor odometry/mapping performance.

%% file: limitations.tex
\section{Conclusion and Future Work}
\name\ creates a lidar-like high resolution point cloud from low resolution single-chip mmWave radar input for use in robotic applications where lidar fails. \name\ designs a machine learning pipeline for this task and overcomes the challenges arising from radar artifacts by choosing design parameters. We show our rich point cloud in a variety of scenes - completely new environments and in occlusions such as smoke. We collect a large dataset of radar-lidar raw data pairs, which is useful for other perception tasks. In the future, we hope to solve other challenges in enabling \name\ to be an invaluable asset in situations where lidar fails. This includes moving beyond 2D and generating 3D point clouds, tackling highly dynamic scenes, and dealing with 3-dimensional mobile platforms (e.g. UAVs). 

\noindent\textbf{Acknowledgements: } \newtext{This research was supported by NSF (2030154, 2106921, 2007786, 1942902, 2111751), Bosch, CONIX Center, DARPA-TRIAD, ARL and CMU-MFI.}